\documentclass[a4paper,twocolumn]{article}

\usepackage[utf8]{inputenc} 
\usepackage[T1]{fontenc}    
\usepackage{url}            
\usepackage{booktabs}       
\usepackage{amsfonts}       
\usepackage{nicefrac}       
\usepackage{microtype}      
\usepackage[table,xcdraw,dvipsnames]{xcolor}

\usepackage[backref=page]{hyperref}
\usepackage[inline]{enumitem}

\newcommand{\leo}[1]{\textcolor{orange}{leo: #1}}
\newcommand{\anh}[1]{\textcolor{blue}{anh: #1}}
\newcommand{\alex}[1]{\textcolor{teal}{alex: #1}}
\newcommand{\ravi}[1]{\textcolor{purple}{ravi: #1}}
\newcommand{\thomas}[1]{\textcolor{red}{thomas: #1}}

\newcommand{\GuildlineMode}[1]{
  \ifthenelse{\equal{#1}{0}}{
  \renewcommand{\gl}[1]{}
   \renewcommand{\leo}[1]{}
  \renewcommand{\anh}[1]{}
   \renewcommand{\alex}[1]{}
   \renewcommand{\ravi}[1]{}
   \renewcommand{\thomas}[1]{}
  }{}
}
\GuildlineMode{1}
\usepackage{float}
\usepackage{graphicx}
\usepackage{subfig}
\usepackage{mathtools}
\usepackage{multirow}
\usepackage{multicol}
\usepackage{tabularx}
\usepackage{lipsum}
\usepackage{ragged2e}
\usepackage{epsfig}
\usepackage{calc}
\usepackage{amssymb}
\usepackage{amstext}
\usepackage{amsmath}
\usepackage{amsthm}
\usepackage{multicol}
\usepackage{pslatex}
\usepackage{apalike}
\usepackage{SCITEPRESS}     

\begin{document}

\title{LiDAR dataset distillation within bayesian active learning framework\\ \large Understanding the effect of data augmentation }

\author{\authorname{Ngoc Phuong Anh Duong\sup{1}, Alexandre Almin \sup{1} Léo Lemarié\sup{1} and B Ravi Kiran \sup{1} \orcidAuthor{0000-0002-8641-7530}}
\affiliation{\sup{1}Machine Learning, Navya, France}
\email{\{first\_name.second\_name\}@navya.tech}
}


\abstract{Autonomous driving (AD) datasets have progressively grown in size in the past few years to enable better deep representation learning. Active learning (AL) has re-gained attention recently to address reduction of annotation costs and dataset size. AL has remained relatively unexplored for AD datasets, especially on point cloud data from LiDARs. This paper performs a principled evaluation of AL based dataset distillation on (1/4th) of the large Semantic-KITTI dataset. Further on, the gains in model performance due to data augmentation (DA) are demonstrated across different subsets of the AL loop. We also demonstrate how DA improves the selection of informative samples to annotate.  We observe that data augmentation achieves full dataset accuracy using only 60\% of samples from the selected dataset configuration. This provides faster training time and subsequent gains in annotation costs.}

\onecolumn \maketitle \normalsize \setcounter{footnote}{0} \vfill

\section{Introduction} 
Autonomous driving has witnessed a recent increase in research and industry-based large-scale datasets in the point cloud domain such as Semantic-KITTI \cite{behley2019semantickitti} and Nuscenes \cite{caesar2020nuscenes}. These datasets enable diverse driving scenarios and lighting conditions, along with variation in the poses of on-road obstacles. The collection procedure
frequently involves recording temporal segments with key frames that are manually selected.
However, these large-scale point clouds datasets have high redundancy, especially in training Deep Neural Network (DNN) architectures. This is mainly due to the temporal correlation between point clouds scans, the similar urban environments and the symmetries in the driving environment (driving in opposite directions at the same location). Hence, data redundancy can be seen as the similarity between any pair of point clouds resulting from geometric transformations as a consequence of ego-vehicle movement along with changes in the environment. Data augmentations (DA) are transformations on the input samples that enable DNNs to learn invariances and/or equivariances to said transformations \cite{anselmi2016invariance}. DA  provides a natural way to model the geometric transformations to point clouds in large-scale datasets due to ego-motion of the vehicle.

Active Learning (AL) is an established field that aims at interactively annotating unlabeled samples guided by a human expert in the loop. With existing large datasets, AL could be used to find a core-subset with equivalent performance w.r.t a full dataset. This involves iteratively selecting subsets of the dataset that greedily maximises model performance. As a consequence, AL helps reduce annotation costs, while preserving high accuracy. AL distills an existing dataset to a smaller subset, thus enabling faster training times in production. It uses uncertainty scores obtained from predictions of a model or an ensemble to select informative new samples to be annotated by a human oracle. Uncertainty-based sampling is a well-established component of AL frameworks today \cite{settles2009active}. 

This paper studies the dataset distillation or reduction of redundant samples on point clouds from the Semantic-KITTI dataset. We note that Semantic-KITTI with 23201 point cloud samples was generated by continuous motion of the ego-vehicle in urban environments in Germany. After testing different options to evaluate uncertainty, we show that DA techniques, if carefully chosen and applied, can improve the selection of informative samples in an AL pipeline. Contributions of the current study include:
\begin{enumerate}
    \itemsep0pt
    \item An evaluation of Bayesian AL methods on a large point cloud dataset for semantic segmentation.
    \item Evaluating existing heuristic function, BALD  \cite{houlsby2011bayesian} for the semantic segmentation task within a standardized AL library \cite{atighehchian2019baal}\cite{atighehchian2020bayesian}. The BALD heuristic used in conjunction with DA techniques shows a high labeling efficiency on a 6000 sample subset of the Semantic-KITTI dataset.
    \item Key ablation studies on informativeness of dataset samples vs data augmented samples that reflect how DA affects the quality of AL based sampling/acquisition function.
    \item A competitive compression over the baseline accuracy while using only 60\% of the dataset under study.
\end{enumerate}
Like many previous studies on AL, we do not explicitly quantify the amount of redundancy in the datasets and purely determine the trade-off of model performance with smaller subsets w.r.t the original dataset.

\section{Related work}

The reader can find details on the major approaches to AL in the following articles: uncertainty-based approaches \cite{gal2017deep}, diversity-based approaches \cite{sener2018active}, and a combination of the two \cite{kirsch2019batchbald}\cite{ash2020deep}. Most of these studies were aimed at classification tasks. Adapting diversity-based frameworks usually applied to a classification, such as \cite{sener2018active}, \cite{kirsch2019batchbald}, \cite{ash2020deep}, to the point cloud semantic segmentation task is computationally costly. This is due to the dense output tensor from DNNs with a class probability vector per pixel, while the output for the classification task is a single class probability vector per image.
Various authors in \cite{kendall2017uncertainties}\cite{golestaneh2020importance}, Camvid \cite{Brostow2009SemanticOC} and Cityscapes\cite{cordts2016cityscapes} propose uncertainty-based methods for image and video segmentation. 
However, very few AL studies are conducted for point cloud semantic segmentation. Authors \cite{wu2021redal} evaluate uncertainty and diversity-based approaches for point cloud semantic segmentation. This study is the closest to our current work.

Authors \cite{birodkar2019semantic} demonstrate the existence of redundancy in CIFAR-10 and ImageNet datasets, using agglomerative clustering in a semantic space to find redundant groups of samples. As shown by \cite{chitta2019training}, techniques like ensemble active learning can reduce data redundancy significantly on image classification tasks. Authors \cite{beck2021effective} show that diversity-based methods are more robust compared to standalone uncertainty methods against highly redundant data. Though authors suggest that with the use of DA, there is no significant advantage of diversity over uncertainty sampling. Nevertheless, the uncertainty was not quantified in the original studied datasets, but were artificially added through sample duplication. This does not represent real word correlation between sample images or point clouds. Authors \cite{hong2020deep} uses DA techniques while adding the consistency loss within a semi-supervised learning setup for image classification task. 

\section{Method}

In this section, we describe our setup used to evaluate the performances of AL for point cloud semantic segmentation, including dataset setup, DNN model architecture, the chosen DA techniques, and most importantly the setup on our AL experiments.

\paragraph{Dataset}
Although there are many datasets for image semantic segmentation, 
few are dedicated to point clouds.
The Semantic-KITTI dataset \& benchmark by authors \cite{behley2019semantickitti} provides more than 43000 point clouds of 22 annotated sequences, acquired with a Velodyne HDL-64 LiDAR. Semantic-KITTI is by far the most extensive dataset with sequential information. All available annotated point clouds, from sequences 00 to 10, for a total of 23201 point clouds, are later randomly sampled, and used for our experiments. 

\begin{figure*}[ht]
    \centering
    \includegraphics[width=0.8\textwidth]{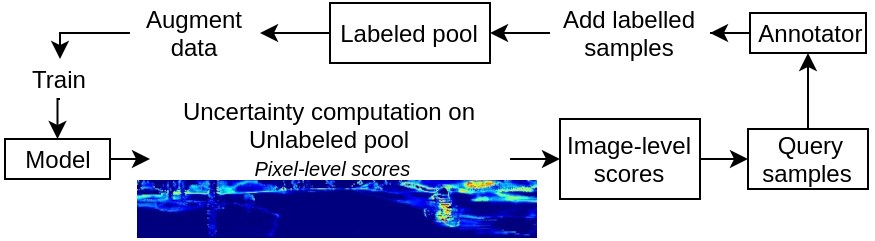}
    \caption{Global flow of active learning on range images from point clouds using uncertainty methods.}
    \label{fig:al_range_image}
\end{figure*}

\paragraph{Model}
Among different deep learning models available, we choose SqueezeSegV2 \cite{wu2018squeezesegv2}, a spherical-projection-based semantic segmentation model, which performs well with a fast inference speed compared to other architectures, thus reduces training and uncertainty computation time. We apply spherical projection \cite{wu2018squeezesegv2} on point clouds to obtain a 2D range image as an input for the network shown in figure \ref{fig:al_range_image}. To simulate Monte Carlo (MC) sampling for uncertainty estimation \cite{gal2016dropout}, a 2D Dropout layer is added right before the last convolutional layer of SqueezeSegV2 \cite{wu2018squeezesegv2} with a probability of 0.2 and turned on at test time. 

\paragraph{Spherical projection}

Rangenet++ architectures by authors \cite{milioto2019rangenet++} use range image based spherical coordinate representations of point clouds to enable the use of 2D-convolution kernels. The relationship between range image and LiDAR coordinates is the following:
\begin{equation*}
        \begin{pmatrix} 
        u \\
        v \end{pmatrix} =  
            \begin{pmatrix} 
                \frac{1}{2} [1-\arctan(y,x)\pi^{-1}]\times w \\
                [1-(\arcsin(z\times r^{-1})+{f}_{up})\times {f}^{-1}]\times h 
            \end{pmatrix} ,  
\end{equation*}
 where $(u,v)$ are image coordinates, $(h,w)$ the height and width of the desired range image,  $f =f_{up}+f_{down}$, is the vertical $fov$ of the sensor, and $r = \sqrt{x^2 + y^2 + z^2}$, range measurement of  each  point. The input to the DNNs used in our study are images of size $W \times H \times 4$, with spatial dimensions $W, H$ determined by the FOV and angular resolution, and 4 channels containing the $x, y$ coordinates of points, $r$ range or depth to each point, $i$ intensity or remission value for each point.
 
\paragraph{Bayesian uncertainty-based approach of AL}

In a supervised learning setup, given a dataset $\mathcal{D}:= \{(\mathbf{x}_1, y_1), (\mathbf{x}_2, y_2), \ldots, (\mathbf{x}_N, y_N) \} \subset \mathcal{X} \times \mathcal{Y}$, 
the DNN is seen as a high dimensional function $f_{\omega} : \mathcal{X} \to \mathcal{Y}$ with
model parameters $\omega$. A simple classifier maps each input $x$ to outcomes $y$. A good 
classifier minimizes the empirical risk $l : \mathcal{Y}\times \mathcal{Y} \to
\mathbb{R}$, which is defined with the expectation $R_\text{emp} (f):= \mathbb{P}_{X, Y} [Y \neq f(X)] $.
The optimal classifier is one that minimizes the above risk.
Thus, the classifier's loss does not explicitly refer to sample-wise uncertainty but
rather to obtain a function which makes good predictions on average. 

Predictive uncertainty \cite{hullermeier2021aleatoric} estimates uncertainty over each  
prediction $\hat{y} = f_{\omega}(\mathbf{x}) = p(y|\mathbf{x})$ given its input $\mathbf{x}$.
A model's predictive uncertainty is a combination of the \emph{aleatoric uncertainty}, irreducible uncertainty due to intrinsic randomness of underlying process, and the \emph{epistemic uncertainty},  reducible uncertainty caused due to missing knowledge, and could be reduced given additional information.

Authors \cite{gal2016dropout} propose generation of MC samples for a given model and input, by activating standard dropout layers at inference time. This provides an uncertainty estimation by sampling different values of DNN weights. Readers can consult work by \cite{gawlikowski2021survey} for uncertainty estimation in DNNs.

\paragraph{Key components of AL framework } We shall use the following terminologies to describe our AL training setup.
\begin{enumerate}
\item \textit{Labeled dataset} $D = \{ \left( \mathbf{x}_i, y_i \right) \}_{i=1}^N$ where $\mathbf{x}_i \in W \times H \times 4$ are range images with 4 input channels, $W, H$ are spatial dimensions, and $y_i \in W \times H \times C$  are one-hot encoded ground truth with $C$ classes. The output of the DNN model is distinguished from the ground truth as $\hat{y}_i$ with the same dimensions.
\item \textit{Labeled pool} $L \subset D$ and a unlabeled pool $U \subset D$ considered as a data with/without any ground-truth, where at any AL-step $L \cup U = D$, the subsets are disjoint and restore the full dataset. 
\item Query size $B$, also called a \textit{budget}, to fix the number of unlabeled samples selected for labeling 
\item  Acquisition function, known as heuristic, providing a score for each pixel given the output $\hat{y}_i$ of the DNN model, $f: \mathbb{R}^{W \times H \times C} \to \mathbb{R}^{W \times H}$ 
\item Including the usage of MC iterations the output of the DNN model could provide several outputs given the same model and input, $\hat{y}_i \in W \times H \times C \times T$ where $T$ refers to the number of MC iterations.
\item \textit{Subset model} $f_L$ is the model trained on labeled subset $L$
\item \textit{Aggregation function} $a: \mathbb{R}^{W \times H \times C \times T} \to \mathbb{R}^+$ is a function that aggregates heuristic scores across all pixels in the input image into a positive scalar value, which is used to rank samples in the unlabeled pool.
\end{enumerate}

\paragraph{Heuristic} Heuristic functions are transformations over the model output probabilities $p(y|x)$ that define uncertainty-based metrics to rank and select informative examples from the unlabeled pool at each AL-step. We used the following uncertainty-based metrics in our experiments: 

    1. \textit{Certainty} heuristic measures the least confident class probability across the highest confident prediction over different MC iterations : $$\text{min}_{y} \text{max}_i 
    \{ f_{\omega}(\mathbf{x}) \}_{i=1}^T $$ where $T$ is the number of MC iterations. 
    
    2. \textit{Entropy} heuristic measures the entropy over predicted class probabilities
    $$H(y|x, L) = -\sum_{c}^{m}p(y=c|x, L) log(p(y=c|x, L))$$
    
    3. \textit{Variance} computes the variance of predictions from model parameters for each class, then averages all variances from all classes to obtain the aggregated score for a sample in classification, or a pixel in image semantic segmentation. The heuristic selects the samples having the highest aggregated scores. The variance for each class 
    $\sigma^2(p(y=c|x, L))$ is: 
     \begin{equation} 
    \frac{1}{T}\sum_{i=1}^{T}(p(y=c|x, w_{i}|L) - p(y=c|x, L))^2
    \label{heuristic:variance}
    \end{equation}

    4. \textit{BALD} \cite{houlsby2011bayesian} selects samples maximizing information gain between the predictions from model parameters, using MC Iterations. The expectation in the equation below is performed over model parameters $\omega$. The information gain $I(y, \omega|x, L)$ is given by
    \begin{equation}
     H(y|x, L) - E_{p(\omega| L)}(H(y|x, \omega))   \label{heuristic:BALD} 
    \end{equation}

\begin{figure*}[ht]
    \centering
    \subfloat[Random dropout mask]{\includegraphics[width=0.47\textwidth]{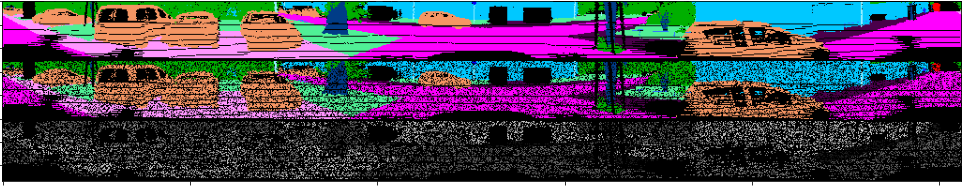}}  \hfill  
    \subfloat[CoarseDropout of Albumentations library]{ \includegraphics[width=0.47\textwidth]{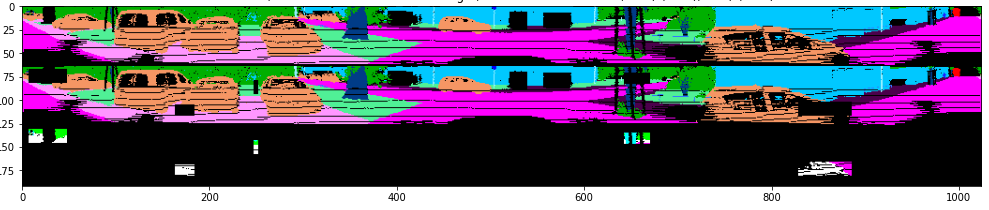}} \\
    \subfloat[Gaussian noise applied on depth channel] { \includegraphics[width=0.47\textwidth]{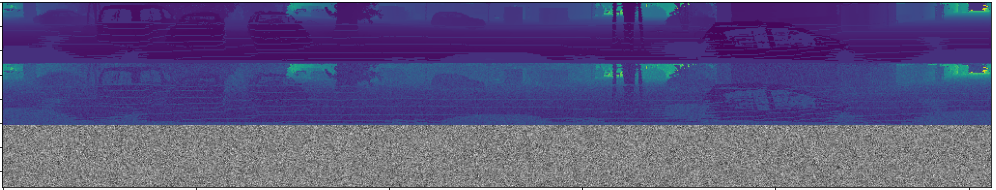}} \hfill  
    \subfloat[Gaussian noise applied on remission channel]{ \includegraphics[width=0.47\textwidth]{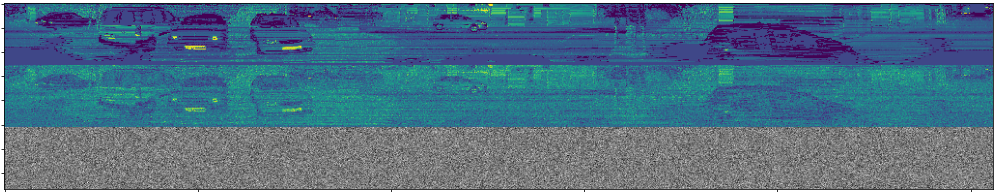}}\\
    \subfloat[Random cyclic shift range image ]{\includegraphics[width=0.47\textwidth]{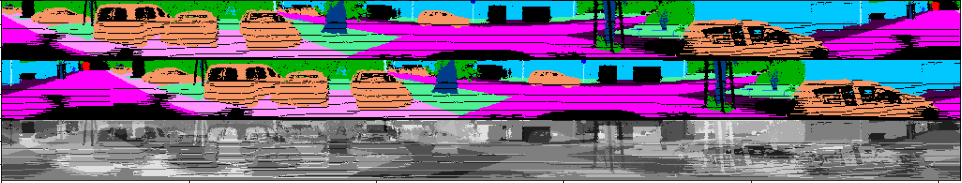}}    \\
    
    \subfloat[Instance Cut Paste]{ \includegraphics[width=0.99\textwidth]{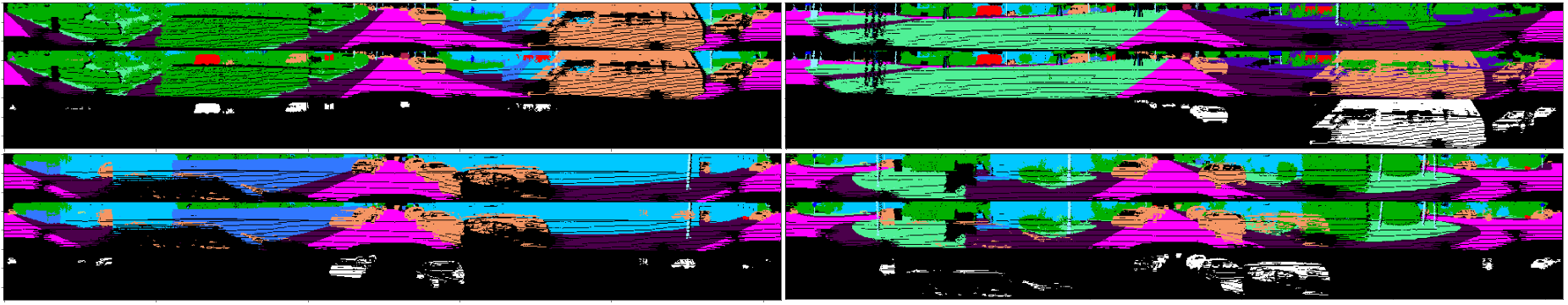}}    
    
    \caption{Before and after applying transformations on Semantic-KITTI. Each image corresponds to a sample such that inner images, from top to bottom, are before and after applying transformations, and the error between them. a, b, c, d are directly used or customized based on Albumentations library \cite{info11020125}}
    \label{fig:da}
       
\end{figure*}

\paragraph{Data augmentations on range images}
We apply DA directly on the range image projection. We selected known effective transformations: (a) \textit{Random dropout mask} on range image and its target by creating a binary mask with uniform dropout probability $p \in [0.1, 0.5]$; (b) \textit{CoarseDropout} which randomly masks out rectangular regions by applying with the following parameters:  max\_height: 16, max\_holes: 5, max\_width: 64, min\_height: 1, min\_holes: 2, min\_width: 1; (c) \textit{Gaussian noise} on depth of range image by with the following parameters $\mu=0 , \sigma^2 \in [0.05, 0.1]$ ; (d) \textit{Gaussian noise} on remission channel of range image with the following parameters: $\mu=0, \sigma^2 \in [0.5, 1.0]$; (e) \textit{Random cyclic shift} on range image (corresponding to rotations on point cloud) and its target to left and right, from 0 to 22.5 degrees around the center; (f) \textit{Instance Cut Paste} randomly copying and pasting instances from one scan to another within a batch. More description and experiment setup of these transformations are in figure \ref{fig:da}.

\begin{table*}[]
\centering
\medskip
\resizebox{\textwidth}{!}{
\captionsetup{justification=centering}
\begin{tabular}{|cccccccc|}
\hline
\multicolumn{3}{|c||}{\cellcolor[HTML]{ECF4FF}Data related parameters} & \multicolumn{5}{c|}{\cellcolor[HTML]{ECF4FF}AL Hyper parameters} \\\hline
\textbf{Range image resolution} & \textbf{Total pool size} &
 \multicolumn{1}{c||}{\textbf{Test pool size}}  & \textbf{Init set size} & \textbf{Budget} & \textbf{MC Dropout} & \textbf{AL steps} & \textbf{Aggregation} \\
1024x64 & 6000 &  \multicolumn{1}{c||}{2000} & 240 & 240 & 0.2 & 25 & sum \\\hline
\multicolumn{8}{|c|}{\cellcolor[HTML]{ECF4FF}Hyper parameters for each AL step} \\\hline
\multirow{2}{*}{\textbf{Max train iterations}} & \multirow{2}{*}{\textbf{Learning rate (LR)}} & \multirow{2}{*}{\textbf{LR decay}} & \multirow{2}{*}{\textbf{Weight decay}} & \multirow{2}{*}{\textbf{Batch size}} & \multicolumn{3}{||c|}{\textbf{Early stopping}} \\\cline{6-8}
 &  &  &  &  & \multicolumn{1}{||c}{\textbf{Evaluation period}} & \textbf{Metric} & \textbf{Patience} \\
100000 & 0.01 & 0.99 & 0.0001 & 16 & \multicolumn{1}{||c}{500} & train mIoU & 15\\
\hline
\end{tabular}
}
\caption{Common experiments settings to each active learning (AL) run.}
\label{tab:setup}
\end{table*}

\paragraph{Evaluation metrics} To evaluate the performance of our experiments we are using the following metrics:

\vspace{1em}

1. \textit{MeanIoU} Intersection over Union (IoU) \cite{song2016semantic}, known as Jaccard index, measures the number of common pixels between the target and prediction masks over the total number of pixels. MeanIoU (mIoU) is mean value of IoU over all classes. Given $TP_c$, $FP_c$, and $FN_c$ as the number of true positive, false positive, and false negative predictions for class c, and C is the number of classes, MeanIoU can be formulated as
    \begin{equation}
    \frac{1}{C}\sum^{C}_{c=1} \frac{TP_c}{TP_c+FP_c+FN_c} 
    \label{eqn:mIoU}
    \end{equation}

\vspace{1em}

2. \textit{Labeling efficiency} Authors \cite{beck2021effective} use the labeling efficiency (LE) to compare the amount of data needed among different sampling techniques with respect to a baseline. In our experiments, instead of accuracy, we use MeanIoU as the performance metric. Given a specific value of MeanIoU, the labeling efficiency is the ratio between the number of labeled range images, acquired by the baseline sampling and the other sampling techniques.
    \begin{equation}
    \text{LE} = \frac{n_{\text{labeled}\_\text{others}}(\text{MeanIoU}=a)}{n_{\text{labeled}\_\text{baseline}}(\text{MeanIoU}=a)}    
    \end{equation}
The baseline method is usually the random heuristic.



\paragraph{Experimental Setup}

As seen in figure \ref{fig:al_range_image} we follow a Bayesian AL using MC Dropout. The heuristic computes uncertainty scores for each pixel. To obtain the final score per range image, we use \emph{sum} as an aggregation function to combine all pixel-wise scores of an image into a single score. At each AL step, the unlabeled pool is ranked w.r.t the aggregated score. A new query of samples limited to the budget size is selected from the ranked unlabeled pool. The total number of AL steps is indirectly defined by budget size, $n_{AL} = \left| D \right|/B$

Based on this pipeline, we made AL runs across different heuristics: random, BALD \cite{houlsby2011bayesian}, entropy and certainty, with and without the application of  DA applied during training time. As mentioned in Table \ref{tab:setup}, we only use 6000 randomly chosen samples from Semantic-KITTI over the 23201 samples available, because every experiment is very time-consuming, and our resources were limited. At each training step, we reset model weights to avoid biases in the predictions, as proven by \cite{beck2021effective}.

In order to evaluate the performances of our pipeline over each experiment, on test set we use LE and MeanIoU as our metrics. Finally, to speed up the training steps, we use early stopping based on the stability of training MeanIoU over $\textit{patience}  * \textit{evaluation}\_\textit{period}$ iterations.

\section{Experiments}
Based on previously described AL configurations, we investigate (A) which heuristic performs the best on semantic segmentation for point clouds, (B) the impact of DA techniques on LE, (C) the informativeness of dataset vs data augmented samples across AL steps and (D) finally the DNN model's stability for sample selection.

\begin{figure}[h]
    \centering
    \includegraphics[width=0.8\columnwidth]{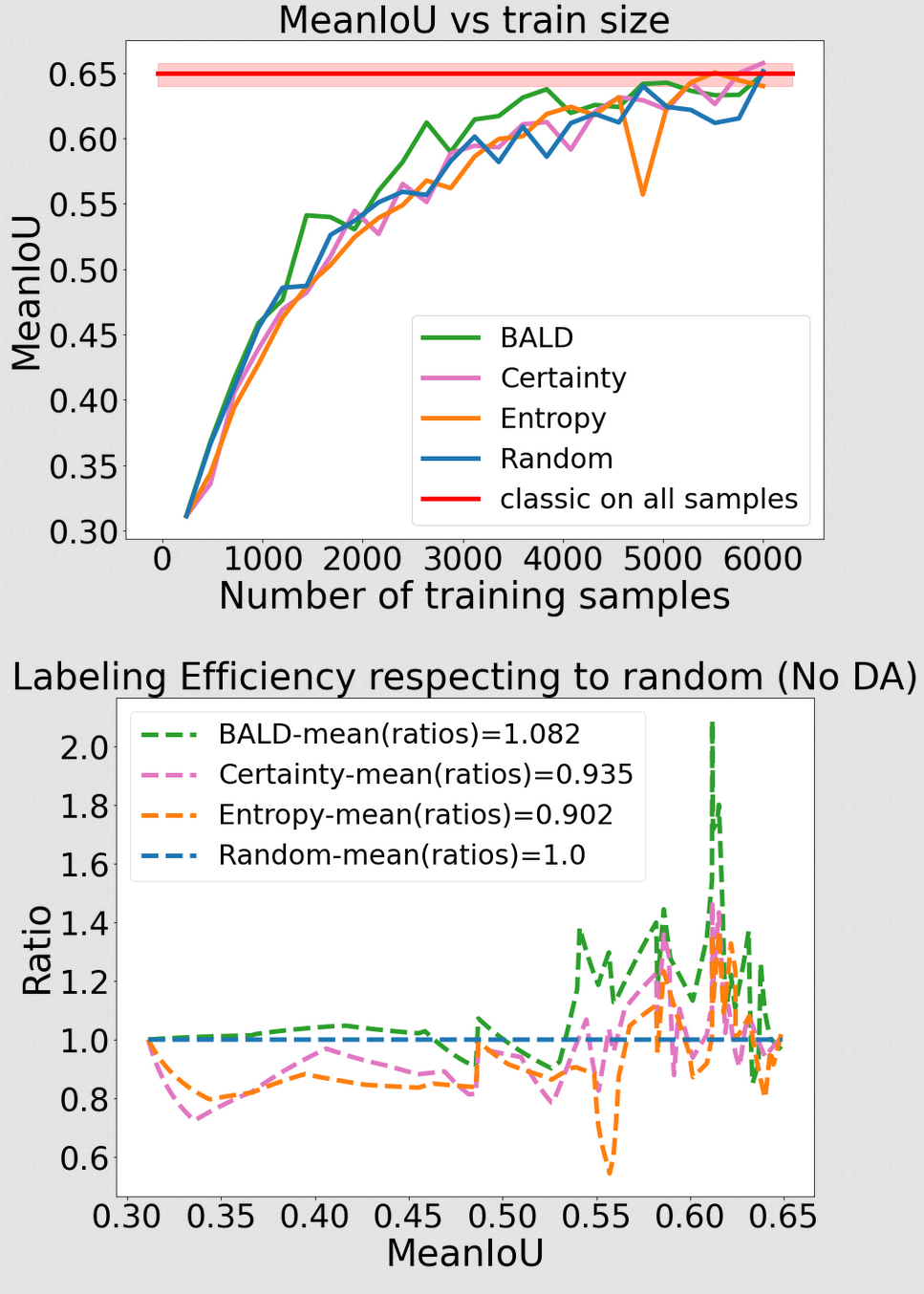}
    \caption{MeanIoU vs number of training samples and labeling efficiency evaluated on test set. Using 100\% of available samples at the end of each run allows us to define an average top performance.}
    \label{fig:experiment:no_da_heuristics}
\end{figure}

\paragraph{A. Heuristic performances on dataset compression}
Firstly, we evaluated the performances of the random heuristic, which is our baseline method. Each complete AL run can achieve the goal performance using fewer number of labeled samples (Figure \ref{fig:experiment:no_da_heuristics}). BALD outperforms other heuristics, allowing the model to converge faster with the highest LE ratio.
To focus on our study on DA, we restrict our focus to the BALD and random heuristics for the rest of the experiments.

\begin{figure*}[h]
    \centering
    \includegraphics[width=0.8\textwidth]{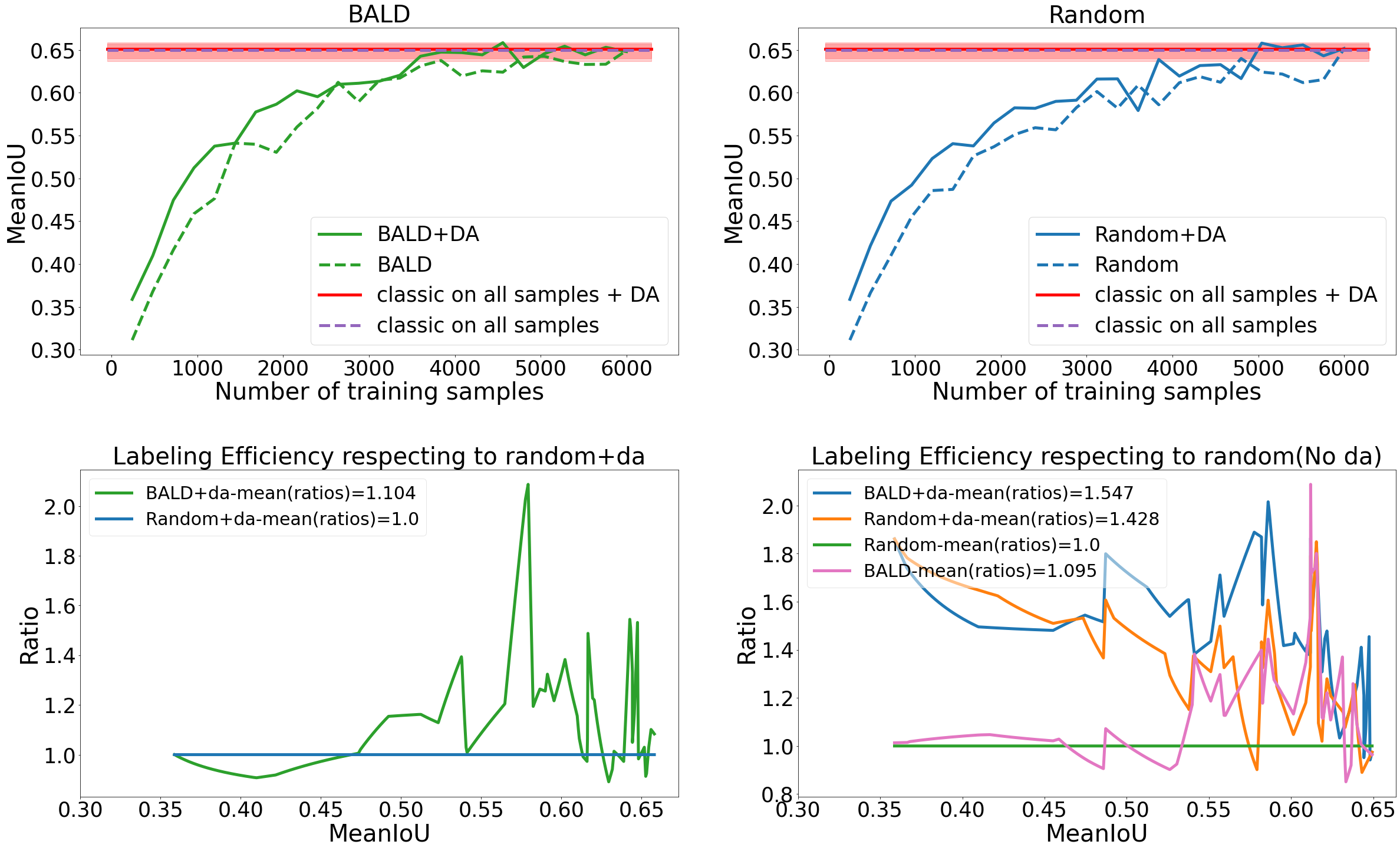}
    \caption{MeanIoU vs number of training samples and labeling efficiency evaluated on test set.}
    \label{fig:experiment:da}
\end{figure*}

\paragraph{B. Data augmentation for dataset distillation}  
In this experiment, DA techniques are applied at training time. On both random and BALD heuristics, figure \ref{fig:experiment:da} shows that DA helps the model to reach the baseline accuracy on test set faster compared to runs without DA. DA provides a significant improvement in the early AL-steps when the subset size are small, as one expects.

DA provides better generalization by regularization. Model output probabilities are 
confident for dataset samples that are similar to DA samples from the labeled pool. In 
other words, with DA, the model tends to select samples different from the trained 
samples and their transformations, and thus reduces redundancy.
BALD with DA can achieve an important dataset distillation, by using only 60\% of the 
total sample pool and still achieving baseline accuracy.

\begin{figure}[h]
    \centering
    \includegraphics[width=\columnwidth]{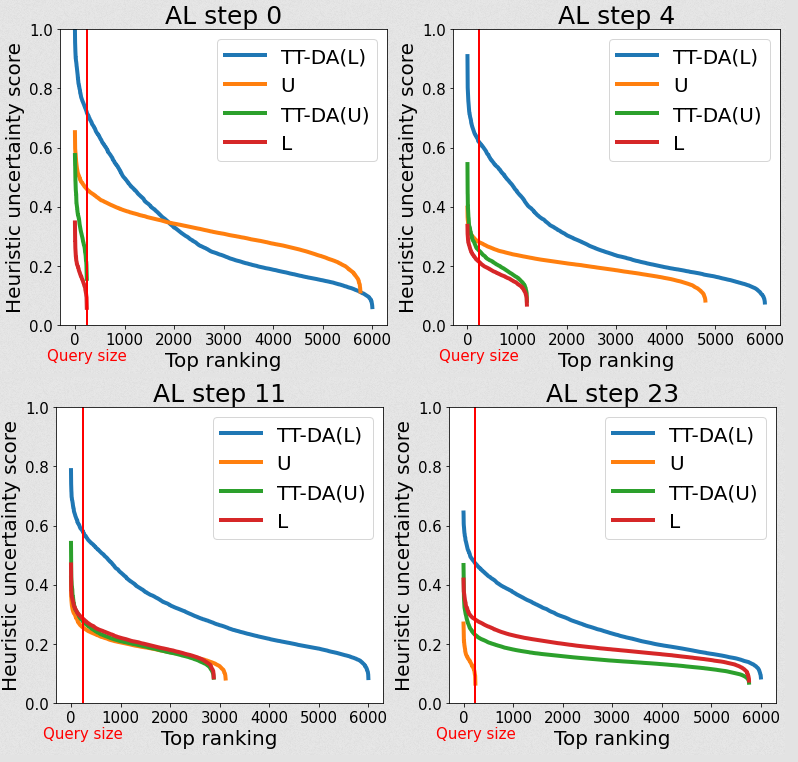}
    \caption{Aggregated heuristic score of samples sorted by decreasing value.}
    \label{fig:experiment:topk}
\end{figure}

\paragraph{C. Heuristic evaluation on augmented samples}
In an effort to understand how data augmented samples affect the heuristic function we evaluated the heuristic function using models trained without DA while predicting on test time augmented images.
We evaluated the aggregated heuristic scores for BALD over firstly the labeled and unlabeled pools, secondly we use Test-Time Data Augmentations (TT-DA) on both labeled and unlabeled pool samples (Figure \ref{fig:experiment:topk}) at different AL steps. To be clear, we used models with no DA during training for this experiment. \emph{(TT-DA(L))} is generated by applying DA at test time on the Labeled pool \emph{(L)} at each training epoch. \emph{(TT-DA(U))} contains augmented samples from the Unlabeled pool \emph{(U)}. We ensure that the combined sizes of \emph{(TT-DA(U))} and \emph{(U)} is always equal to 6000 samples. In figure \ref{fig:experiment:topk} the sorted aggregated scores $a$ to the left of the red line which defines the budget of each AL-step, we notice the following ordering :
These results show that in the early AL steps: $$a(TTDA(L)) > a((U)) > a(TTDA(U)) > a((L))$$ and in final AL steps:
$$a(TTDA(L)) > a(TTDA(U)) > a((L)) > a((U))$$

We observe that during the initial AL step, the aggregated score is low as expected on \emph{(L)}, which has been used to train the model. Because the model has been trained on only 240 samples from \emph{(L)}, the aggregated score is very high on \emph{(U)}, \emph{(TT-DA(U))} and \emph{(TT-DA(L))}. As the AL step goes on, the aggregated scores are globally decreasing, this can be explained by the growing pool of selected data \emph{(L)} used to train the model. In the final AL step, \emph{(U)} has the smallest uncertainty scores as the model is now well trained and able to correctly generalize on unseen samples. The highest aggregated scores are related to DA samples from the labeled \emph{(TT-DA(L))} and unlabeled \emph{(TT-DA(U))} pool. This could be because the DA is providing transformed samples that are now outside the support of the dataset distribution.

\paragraph{D. Model stability \& effectiveness for sampling}
In this part, we study the model stability, based on the mean variance computed on class probabilities across all MC iterations. We also measure the model sampling effectiveness by computing mean BALD metric. 

\begin{figure}[h] 
    \includegraphics[width=0.95\columnwidth]{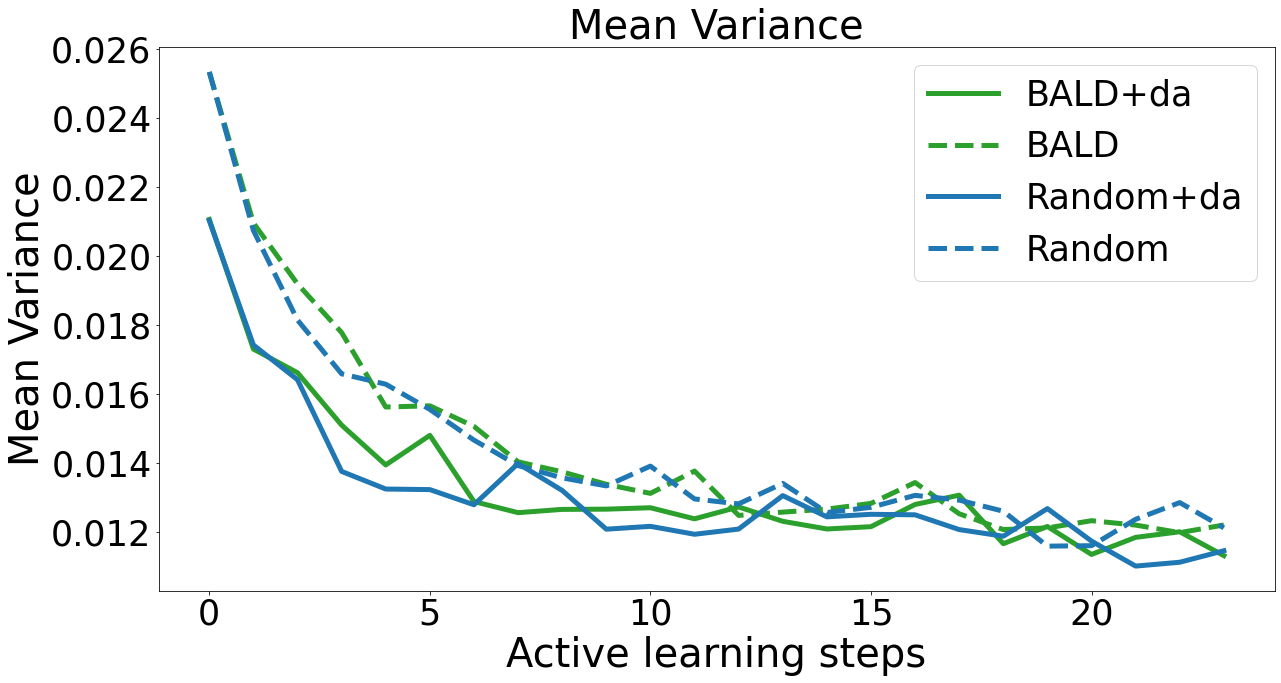}
    \includegraphics[width=0.95\columnwidth]{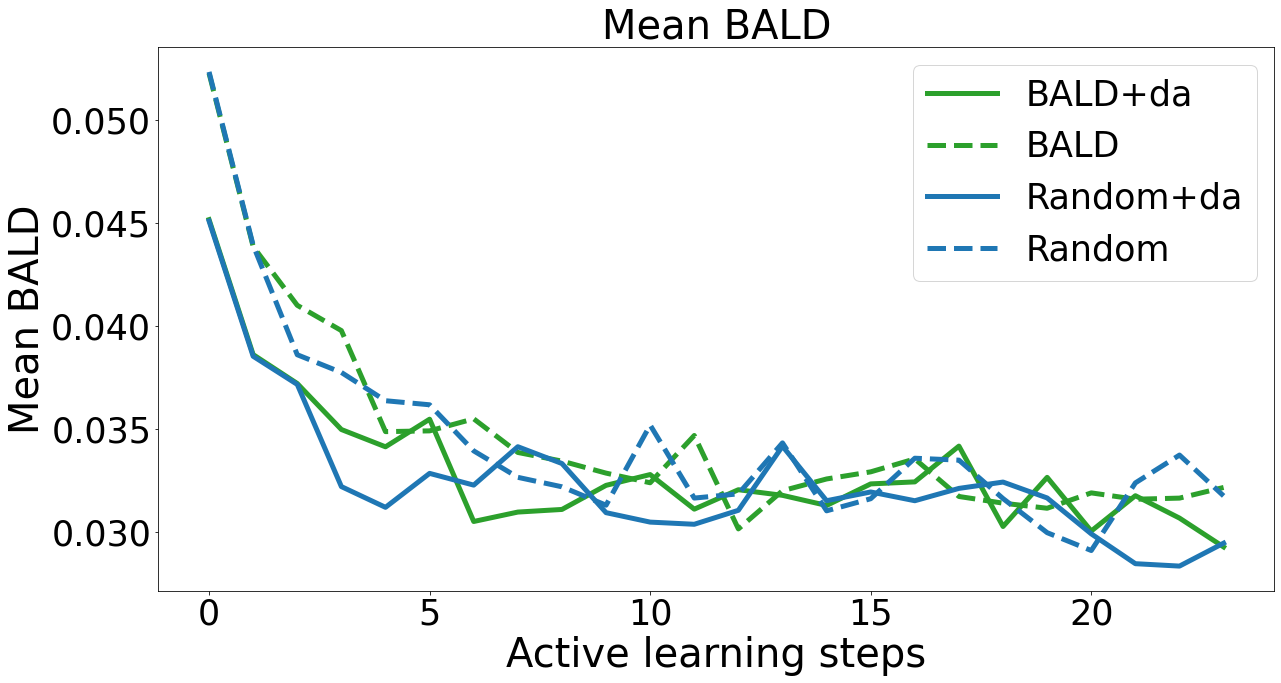}
    \caption{Mean of variances computed on class probabilities over MC predictions (equation \ref{heuristic:variance}) and mean of BALD (equation \ref{heuristic:BALD}) for all pixels across all samples in the test set.}
    \label{fig:experiment:uncertainty_scores}
\end{figure}

Across all AL steps (Figure \ref{fig:experiment:uncertainty_scores}), models with DA 
become confident in earlier AL steps (on account of dropout), and are able to select 
samples that maximise information gain sooner w.r.t models without DA.
This experiment shows that DA improves the stability of models and allows a better and 
faster sample selection by reducing the uncertainty of the heuristic functions.

\paragraph{Key future challenges }
Quantifying data redundancy shall be investigated in our future study based on work by authors in \cite{birodkar2019semantic}, \cite{2021AdversarialAL}. To improve the heuristic function, recent work \cite{lakshminarayanan2016simple} \cite{allingham2021sparse} on explicit ensembles shows strong results for uncertainty computing, and \cite{aghdam2019active} show that adding temporal reasoning can be beneficial for data selection on object detection task. We aim to further our study by experiment on different budget sizes, while testing on the complete Semantic-KITTI dataset. Another key issue in industrial datasets is the filtering or exclusion of corrupted or outlier images/pointclouds from the AL loop \cite{chitta2019training} that frequently affect the ranking of unlabeled pool samples.
Finally, the last but critical step in AL is the stopping criterion to terminate any future input to the AL-pipeline, though this is highly task and dataset dependent. A simple rule can be thresholding the change in entropy over model's output class probabilities. It could also be the incremental gain in the performance of
the model. A much more practical limit would be the budget of human effort allocated to annotation.

\section{Conclusion}
\vspace{-0.9cm}
Our work demonstrates the benefits of data augmentation in active learning (AL) for point cloud semantic segmentation task. It conforms with results by \cite{beck2021effective} for the image classification task on CIFAR dataset. We observe that the effect of data augmentation on BALD heuristic provides a robust and efficient heuristic for sample selection. It not only selects more uncertain samples at each AL step, but also improves the heuristic function's stability, subsequently leading to improved label efficiency. With only 60\% of the samples, we reach the same accuracy as a supervised training with the full selected subset. The computing time gained by training the model on the AL-selected subset from AL w.r.t training on the original dataset could help gain few days to weeks. Thus data augmentation within AL frameworks have helped in reducing annotation costs as well as reducing training time in production over large datasets.

\vspace{-1em}
\paragraph{Acknowledgements}
{\small This work was granted access to HPC resources of [TGCC/CINES/IDRIS] under the allocation 2021- [AD011012836] made by GENCI (Grand Equipement National de Calcul Intensif). It is also part of the Deep Learning Segmentation (DLS) project financed by \href{https://www.ademe.fr/}{ADEME}.}

\bibliographystyle{apalike}
{\small
\bibliography{AL_SKITTI_VISAPP2022.bib}
}

\end{document}